\documentclass[journal]{IEEEtran}
\IEEEoverridecommandlockouts
\usepackage{cite}
\usepackage[T1]{fontenc}
\usepackage{amsmath,amssymb,amsfonts}
\usepackage{graphicx}
\usepackage{textcomp}
\usepackage{CJKutf8}
\usepackage{subfigure}
\usepackage{tabularx}
\usepackage{booktabs}
\usepackage{multirow}
\usepackage{xcolor}
\usepackage{wrapfig}
\usepackage{arydshln}
\usepackage{threeparttable}
\usepackage{url}          
\usepackage{nicefrac}      
\usepackage{microtype}      
\usepackage[utf8]{inputenc} 
\usepackage{hyperref}       
\usepackage{algpseudocode}
\usepackage{amsmath}
\usepackage[linesnumbered,ruled]{algorithm2e}
\usepackage{algpseudocode}
\usepackage{nomencl}
\makenomenclature

\def\BibTeX{{\rm B\kern-.05em{\sc i\kern-.025em b}\kern-.08emb
T\kern-.1667em\lower.7ex\hbox{E}\k
ern-.125emX}}

\begin{document}

\title{EM-DARTS: Hierarchical Differentiable Architecture Search for Eye Movement Recognition\\

}

\author{
       Huafeng Qin,
         Hongyu Zhu,
         Xin Jin,
         Xin Yu,
         ~\IEEEmembership{Senior member, IEEE} Mounim~A.~El-Yacoubi, and Shuqiang Yang.
          
\thanks{H. Qin, H. Zhu, X. Jin, and X. Yu are with National Research base of Intelligent Manufacturing Service, Chongqing Micro-Vein Intelligent Technology Co, the School of Computer Science and Information Engineering, Chongqing Technology and Business University, Chongqing 400067, China (e-mail: qinhuafengfeng@163.com, zhuhongyu@ctbu.edu.cn).} 
\thanks{M. A. El-Yacoubi is with SAMOVAR, Telecom SudParis, Institute Polytechnique de Paris, 91120 Palaiseau, France (e-mail: mounim.el\_yacoubi@telecom-sudparis.eu).}
\thanks{Shuqiang Yang is with China University of Mining and Technology, Xuzhou 221116, China (e-mail: joanhn@163.com).}

\thanks{Manuscript received April XX, 2024; revised XXXX XX, 202X. This work was supported in part by the National Natural Science Foundation of China (Grant Nos. 61976030), and the National Natural Science Foundation of China(CSTB2024NSCQ-MSX1118),   Key project of Henan Provincial Department of Science and Technology (Grant No. 222102210301), the
Science Fund for Creative Research Groups of Chongqing Universities (Grant No. CXQT21034, Grant Nos. KJQN201900848 and KJQN201500814). 

(Corresponding author: Xin Yu and Shuqiang Yang.)}}
\maketitle

\begin{abstract}
Eye movement biometrics has received increasing attention thanks to its highly secure identification. Although deep learning (DL) models have {shown success in eye movement recognition, their architectures largely rely on human prior knowledge.} Differentiable Neural Architecture Search (DARTS) automates the manual process of architecture design with high search efficiency. However, DARTS typically stacks multiple cells to form a convolutional network, which limits the diversity of architecture. Furthermore, DARTS generally searches for architectures using shallower networks than those used in the evaluation, creating a significant disparity in architecture depth between the search and evaluation phases. To address this issue, we propose EM-DARTS, a hierarchical differentiable architecture search algorithm to automatically design the DL architecture for eye movement recognition. First, we define a supernet and propose a global and local alternate Neural Architecture Search method to search the optimal architecture alternately with a differentiable neural architecture search. The local search strategy aims to find an optimal architecture for different cells while the global search strategy is responsible for optimizing the architecture of the target network.  To {minimize} redundancy, transfer entropy is proposed to compute the information amount of each layer, {thereby} further {simplifying the network search process}. {Experimental results} on three public datasets demonstrate that the proposed EM-DARTS is capable of producing an optimal architecture that leads to state-of-the-art recognition performance, {Specifically, the recognition models developed using EM-DARTS achieved the lowest EERs of 0.0453 on the GazeBase dataset, 0.0377 on the JuDo1000 dataset, and 0.1385 on the EMglasses dataset.}
\end{abstract}


\section{Introduction}
Biometrics, an increasingly prevalent technology nowadays, uses digital representations of individual biometric data for authentication and identification. Various biometric technologies have been applied in military and government institutions as well as in our everyday lives. 
Static biometric traits such as fingerprints \cite{Peng2022RadioFF}, face \cite{Liu2018ExploringDF}, finger vein\cite{li202fingervein}, and iris\cite{iris}, are widely applied for identification. Nevertheless, the inherent stability of these characteristics is susceptible to spoofing attacks \cite{ruansong,qin2023palmVein,hurongshan}. Furthermore, certain symptoms, such as glaucoma and fingerprint degradation, can make it difficult for instruments to extract high-quality biometric features \cite{Jain2007BiometricRO}. In contrast to static physiological characteristics, behavioral biometric modalities such as eye movement \cite{Qin2024EmMixformerMT}, gait \cite{Xu2023OcclusionAwareHM}, and signatures \cite{Yang2023InSP}, show natural liveness identification to achieve higher security. Among them, eye movement depends largely on user brain activity and extraocular muscle properties, making it impossible to reproduce, and, therefore, a robust form for authentication, resistant to forgery\cite{Holland2011BiometricIV}.
An oculomotor system is thus endowed with notable advantages in supporting liveness detection \cite{Makowski2020BiometricIA} and spoof-resistant continuous authentication \cite{eberz2015preventing}. Moreover, it can be easily and seamlessly integrated with the iris, pupil \cite{Blignaut2013MappingTP}, and other ocular physiological attributes for multimodal recognition. Eye movement biometrics has, as a result, received considerable attention in the past two decades\cite{katsini2020role}.


\begin{figure*}[htbp]
    \centerline{\includegraphics[scale=0.8]{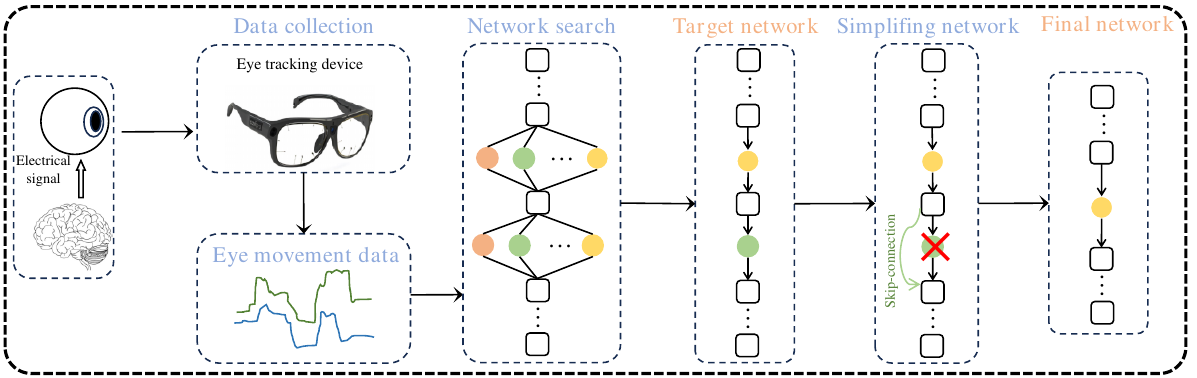}}
    \caption{The overall flow chart of eye movement recognition using NAS algorithm.
    }
    \vspace{-10pt}
    \label{fig1}
\end{figure*}

\subsection{Motivation}
An oculomotor system captures the coordinated contraction and relaxation of the six ocular muscles, regulated by brainstem nerves during visual processing. These movements can be broadly classified into gaze and sweep based on the angular velocity of eye rotation \cite{Holland2012BiometricVV}. Eye tracking devices detect the gaze point's position through near-infrared light reflections on the eye, capturing the trajectory of gaze movements to obtain raw eye movement data.  The resulting data are then subject to feature coding and matching to achieve identification. As eye movements are controlled by neuroelectric signals, the movement data contain a large amount of information about the brain's real-time cognitive processes.  Eye movement biometrics analyzes eye movement patterns, observed during gaze, reading, etc. The extant literature\cite{Akkil2014TraQuMeAT} indicates that eye movement is highly distinctive and stable over time.



The approaches proposed to extract eye movement features can be split into two categories: traditional approaches \cite{Kasprowski2004EyeMI, Kasprowski2005EnhancingEB, Komogortsev2008EyeMP, Komogortsev2010BiometricIV} and DL-based approaches\cite{Makowski2021DeepEyedentificationLiveOB, Lohr2022EyeKY, Taha2023EyeDriveAD, Lohr2021EyeKY, RP-InceptionV3}. The traditional approaches include hand-crafted descriptors, e.g., Fixation Density Map (FDM) \cite{rigas2014biometric} and traditional Machine learning (ML) approaches such as Principal Component Analysis (PCA) \cite{Kasprowski2005EnhancingEB} and {Support Vector Machine} (SVM) \cite{bulling2010eye}. For handcrafted descriptors, researchers assume a specific probability distribution for the data and propose ad hoc models for feature extraction. The complex distribution of human eye movements, however, makes it challenging to create effective mathematical models. Traditional ML approaches avoid explicit feature extraction but still suffer from inadequate recognition performance due to limitations in feature representation capacity. 
By stacking multiple layers to learn, in an end-to-end way, rich robust feature representations from raw images from large training data without any prior assumption, DL-based approaches, by contrast, achieve higher recognition accuracy over traditional methods. Although various DL approaches \cite{Qin2024EmMixformerMT}\cite{Makowski2021DeepEyedentificationLiveOB, Lohr2022EyeKY, Taha2023EyeDriveAD, Lohr2021EyeKY, RP-InceptionV3} have been relatively successfully proposed for eye movement recognition, their model architectures are manually designed by human experts, leading to several issues\cite{ren2021comprehensive}\cite{qin2024AGNAS}: 1) Designing architectures manually requires significant expertise, making it challenging to develop optimal neural networks for specific tasks; 2) Manual designing is a time-consuming and error-prone process; 3) Human experts, due to limited knowledge and experience, may struggle to explore the entire design space of complex deep neural network architectures; 4) Manually designed architectures may not be extended to other tasks.

Neural Architecture Search (NAS) \cite{zoph2018learning} automatically searches for network architectures from a predefined search space. NAS, which has outperformed manually designed architectures on image classification \cite{zoph2018learning} and semantic segmentation \cite{chen2018searching}, has been applied recently in biometrics\cite{qin2024AGNAS}. Current NAS algorithms can be broadly classified into three categories \cite{chang2019data}: reinforcement learning-based NAS\cite{Zoph2016NeuralAS}, evolutionary algorithms-based NAS \cite{Real2018RegularizedEF} and DARTS-based NAS \cite{Liu2018DARTSDA,Liang2019DARTSID,Chu2019FairDE}.  
As searching the network structure in the first two categories is non-differential, it is impossible to optimize the parameters by gradient descent. Instead,  they treated the architecture search as a black-box optimization problem over a discrete domain, which requires a large number of architecture evaluations\cite{Liu2018DARTSDA}. Recently, DARTS, proposed to reduce search cost, has become a mainstream neural architecture search approach. By transforming discrete selection operations such as convolution and skip-connection into a continuous space, allowing architecture search to be optimized via gradient descent based on validation set performance, DARTS has become the state of the art by performing a delicate balance between search cost and performance. DARTS, nonetheless, suffers from well-known performance collapse due to an inevitable aggregation of skip-connections \cite{Chu2019FairDE}. Current DARTS-based methods\cite{Liu2018DARTSDA,Liang2019DARTSID,Chu2019FairDE,Movahedi2022lamda} suffer from the following problems: 1) They define a cell, a Directed Acyclic Graph (DAG) of $N$ nodes, and search for an optimal cell within a predefined network structure, stacking multiple copies of this cell to form the final network for downstream tasks. All cells in the final network share the same architecture parameters, which reduces the diversity and limits representation capacity; 2) The architecture search is guided by a shallow network while a deeper network is used as the final network for the test.  As there is a huge difference between shallow and deeper networks, the resulting architecture is optimal during the search process but not for test.

\subsection{Our Work}
To address these problems, in this paper, we propose  {EM-DARTS, a hierarchical differentiable architecture search algorithm (Fig. \ref{fig1}). It efficiently designs lightweight network structures with strong performance. This makes it well-suited for mobile and wearable devices. The lightweight design reduces hardware requirements and extends battery life, which are key factors for portable applications.}

 {The algorithm also enables large-scale deployment in practical settings. In public safety, it can be used for identity verification, where cost efficiency and fast processing are critical. In commercial applications, such as personalized marketing or behavior tracking, its efficiency lowers operational costs. By minimizing hardware demands, EM-DARTS supports the widespread use of eye movement recognition technology in diverse real-world scenarios.}

This work builds upon prior research \cite{Zhu2024RelaxDR}. First, we propose a local differentiable architecture search strategy to search the architecture of cells at different layers. Then, a global differentiable architecture search strategy is proposed to optimize the architecture of the target neural network. Both strategies are integrated to form a hierarchical differentiable architecture search approach, trained in an alternative way. Finally, a transfer entropy strategy is proposed to optimize the number of layers.

The contributions of our work are summarized below:
\begin{itemize}
    \item We are the first to  {investigate} NAS  {for} eye movement recognition, by proposing EM-DARTS, a differentiable architecture search algorithm to automatically search for the optimal network.  {As the architecture of the neural network is automatically derived by exploring an extensive candidate architecture space, the resulting model demonstrates superior performance compared to manually designed networks. The experimental results (Tables \ref{table2}-\ref{table7}) on three public eye movement databases imply that the NAS-based approaches outperform the manual designing approaches to accuracy improvement.}

    \item We present a hierarchical neural architecture search method, which searches the optimal global and local architecture in an alternative way with a differentiable network architecture. Different from traditional DARTS sharing parameters across all cells, our method independently searches each cell during the local search stage so that the cells in different layers have different architecture parameters, which improves the feature representation capacity of the target neural network. The target neural network architecture is further optimized by alternate training, which avoids creating a gap between the architecture depths in the search and evaluation scenario.  {The experimental results (Fig.8) demonstrate that our neural architecture search method is capable of searching different architecture parameters for different cells.} 
    \item We propose a Transfer Entropy-based Simplification strategy to simplify the target neural network. Specifically, we calculate \textit{transfer entropy} to evaluate the importance of each layer and remove the layers with the lowest transfer entropy through a greedy search approach.  { As the layers of the neural network are refined using this simplification strategy, a more optimal architecture is achieved, leading to improved performance. Ablation experiments demonstrate that the proposed simplification strategy reduces the EER of the model from 0.0609 to 0.0537 on the RAN sub-dataset.}

    \item  {Our comprehensive experiments on three public eye movement datasets demonstrate that our approach establishes a new state-of-the-art performance in the field. Specifically, it achieves EERs of 0.0520, 0.0453, 0.0537, and 0.1481 on four sub-datasets in the GazeBase dataset, 0.0377 on the JuDo1000 dataset, and 0.1358 on the EMglasses dataset.}
\end{itemize}

\begin{figure}[htbp]
\centerline{\includegraphics[scale=0.9]{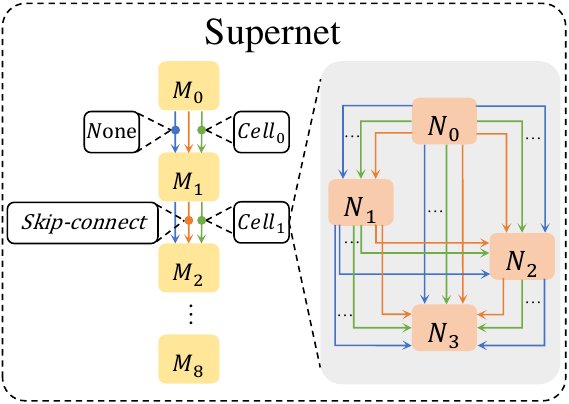}}
    \caption{Architecture of the supernet, where arrows indicate the candidate operations between nodes.}
    \label{supernet}
    \vspace{-10pt}
\end{figure}

\section{Related Work}
\subsection{Traditional Eye Movement  Recognition Method}
Traditional methods use manual techniques to extract useful features from eye movement data and then apply ML algorithms for recognition. Common features include eye movement velocity, gaze duration, and path length.
In 2004, Kasprowski et al.\cite{Kasprowski2004EyeMI} proposed using eye-movement features for identity recognition on a self-constructed dataset. In a later study, \cite{Kasprowski2005EnhancingEB} used PCA for dimensionality reduction of eye movement features, while Komogortsev et al. \cite{Komogortsev2008EyeMP} considered the bioanatomic properties of the eyeball and established a linear horizontal oculomotor plant mechanical model.  \cite{Komogortsev2010BiometricIV, Komogortsev2012BiometricAV} expanded on this foundation and proposed the concept of Oculomotor Plant characteristics (OPC). In \cite{Holland2011BiometricIV}\cite{Holland2013ComplexEM}, Holland et al. extracted average fixation duration, average vectorial saccade amplitude, and more than a dozen more complex eye movement features for identification. In 2015, Kasprowski et al. \cite{Kasprowski2016UsingDM} used the dissimilarity matrix to measure the similarity of eye movement data for classification, while in 2017, Bayat et al. \cite{Bayat2017BiometricIT} conducted recognition experiments on a self-constructed dataset, using eye movement data combined with pupil size. Subsequently, Li et al. \cite{Li2018BiometricRV} extracted features using a multi-channel Gabor wavelet transform. Starting in 2020, researchers began to notice the importance of temporal features in eye movement data, leading to the use of Long Short Term Memory (LSTM) to construct  {Recurrent Neural Network} (RNN) for feature extraction \cite{Yeamkuan2020FixationalFG}.

\subsection{Deep Learning Based Eye Movement Recognition Method} 
In earlier studies, eye movement features were manually extracted and classified using shallow ML models. Owing to their limited representational power, the latter could not effectively learn complex eye movement features. Thanks to their end-to-end feature learning capabilities, DL models are now widely used in eye movement recognition\cite{Qin2024EmMixformerMT} and other tasks\cite{qin2024sumix,jin2024starlknet,LiaoHongchao}. In \cite{Lohr2020AML}, the authors used metric learning for eye movement recognition based on a three-multilayer perceptron for feature encoding after pre-classifying and filtering the raw features. More advanced work, such as Lena et al.'s \cite{Jger2019DeepEB} in 2019, used raw eye movement data for end-to-end recognition. They developed a Convolutional Neural Network (CNN)-based Siamese Network that processes involuntary micro-eye movement data through two separate sub-networks. The studies \cite{Makowski2021DeepEyedentificationLiveOB}\cite{Makowski2022OculomotoricBI} fine-tuned the model and expanded it into the DeepEyedentificationLive (DEL) model. In 2022, Dillon et al.\cite{Lohr2022EyeKY} proposed an exponentially-dilated CNN for recognizing eye movement, to capture more global information; they evaluated it on GazeBase, a large publicly available dataset\cite{Griffith2021GazeBaseAL}. In 2023, Taha et al.\cite{Taha2023EyeDriveAD} collected vehicle driver eye movement data using a remote low-frequency acquisition device. They extracted features by combining LSTM and dense networks for end-to-end driver identification.

 {Recently, several state-of-the-art methods have been proposed. Qin et al. \cite{Qin2024EmMixformerMT} combined temporal and spatial features from eye movement data using an improved LSTM and Transformer architecture. They achieved state-of-the-art performance on both a self-built dataset and public datasets \cite{Makowski2020BiometricIA}\cite{Griffith2021GazeBaseAL}. Peng et al.\cite{RP-InceptionV3} utilized recurrence plot encoding and the InceptionV3 model to analyze and recognize eye movement data. They first transformed the raw eye movement signals into 2-D images through recurrence plot encoding, and then used these images as inputs to the InceptionV3 model for biometric recognition. Furthermore, \cite{gong2024ciabl} introduced a multimodal emotion recognition model CiABL, that leverages both electroencephalogram (EEG) and eye movement signals. This approach provides a novel method for processing eye movement data and expands its potential applications.}

\subsection{Differentiable Architecture Search}
The model's structure significantly impacts performance. In traditional DL methods, designing an effective network structure requires extensive manual testing and evaluation. NAS, by contrast, automates the selection of network structures for optimal performance. Current NAS algorithms follow the basic framework of NASNet\cite{zoph2018learning}, dividing the process into a search space, a search strategy, and a model performance evaluation strategy. The search strategy is the core component, determining the efficiency and effectiveness of the search. Significant progress has been made recently in search strategies, with well-known ones including Evolutionary Algorithms (EA), Reinforcement Learning (RL), and Gradient-based Differentiable (GD) methods. 
Both EA\cite{Real2018RegularizedEF} and RL\cite{Zoph2016NeuralAS} based search strategies treat the search space as discrete, precluding thereby gradient descent for architecture search. DARTS\cite{Liu2018DARTSDA} used \textit{Softmax} to treat the selection of candidate architectures as an optimization problem in a continuous space, enabling gradient-based optimization of architecture weight parameters. This approach is the most popular NAS algorithm currently. In 2019, ProxylessNAS\cite{Cai2018ProxylessNASDN} addressed the inconsistency between the search and evaluation networks of DARTS by using path-level binarization of network structure parameters, enabling the model to learn the architecture for large-scale tasks directly. DARTS+\cite{Liang2019DARTSID} introduced an early stopping mechanism in the search stage to mitigate the issue of skip-connection enrichment, which can lead to significant performance loss in the final model.
Using the \textit{Sigmoid} instead of \textit{Softmax} to score the architectures, Fair DARTS \cite{Chu2019FairDE} also addresses the skip-connection enrichment phenomenon by transforming the candidate operations in the search phase from competition to cooperation. 
P-DARTS\cite{Chen2019ProgressiveDA} reduces the search space and incrementally increases the network layers during the search phase to mitigate the "depth gap" problem. It also mitigates skip-connection enrichment by introducing dropout and limiting the number of skip-connections.
In 2020, Robust DARTS\cite{Zela2019UnderstandingAR} proposed using Hessian feature roots to measure signs of performance collapse in DARTS. The status of the structural parameter $\alpha$ was challenged by BN-NAS\cite{Chen2021BNNASNA} and DARTS-PT\cite{Wang2021RethinkingAS} in 2021, using other indicators for model evaluation. DARTS-\cite{Chu2020DARTS-} proposed using auxiliary skip-connections to leverage the advantages of skip-connections in search operations compared to other operations.
Recent research, $\Lambda$-DARTS, presented in 2023 \cite{Movahedi2022lamda}, showed that the weight sharing framework is flawed, providing an unfair advantage to layers closer to the output in choosing the optimal architecture, leading to performance collapse.
TETLO\cite{Sheth2023ImprovingDN} proposed a three-tier optimization framework aimed at encouraging transferability by improving the main model's network structure and facilitating efficient knowledge transfer to the auxiliary model. CDARTS\cite{Yu2020CyclicDA} established a cyclic feedback mechanism between search and evaluation networks to address the performance gap between them.

\begin{figure*}[htbp]
\centerline{\includegraphics[scale=0.70]{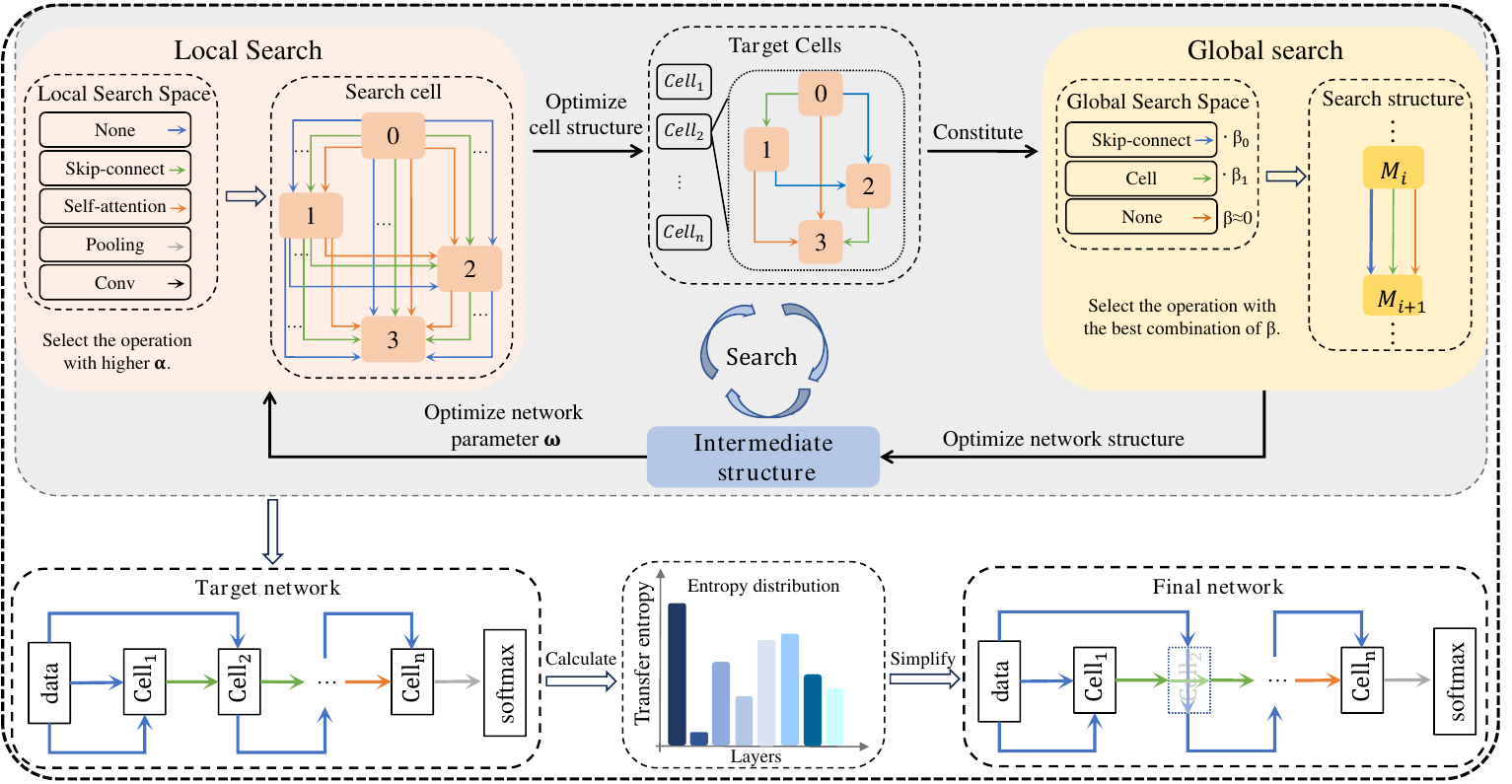}}
    \caption{The framework of the proposed EM-DARTS model.  The local cells and global architecture are alternately optimized to obtain the target network. Then, the resulting target network is simplified by transfer entropy.}
    \label{fig2}
    \vspace{-10pt}
\end{figure*}

\section{Methodology}

In current DARTS-based NAS approaches \cite{Liu2018DARTSDA}\cite{Chu2019FairDE}\cite{Chu2020DARTS-}, architecture search is achieved on a shallow network with 8 cells with the resulting architecture evaluated on a deeper network with 20 cells. As there is a huge difference between the behaviors of shallow and deep networks, the optimal structures during the search process are not optimal for evaluation. In addition, all cells share the same architecture parameters, which results in limited performance. To address these issues, we propose the EM-DARTS algorithm for eye movement recognition. First, we define a  supernet (Fig. \ref{supernet}) and an operation space (Table \ref{table12}). According to the relation between the local cell and the global target network, we divided the operation space into two sub-spaces: global search space and local search space (Table \ref{table12}). A hierarchical NAS method is then proposed to search the optimal architecture from the two spaces alternately with a differentiable network architecture search. To further simplify the network structure, we propose \textit{Transfer Entropy} as an indicator to reduce the layers of the target neural network, so as to reduce redundancy. 


\subsection{Overview EM-DARTS}
NAS is the process of automatically searching the optimal combination of operations in a predefined network search space $\mathcal{O}$. The traditional DARTS-based method aims to search two cells: Normal Cell $\mathcal{N}$ and Reduction Cell $\mathcal{R}$, where $\mathcal{N}$ maintains the output feature dimensions and $\mathcal{R}$ reduces the feature dimensions by half. The resulting cells are stacked to form the final network architecture for different tasks. Differently, we define a supernet and propose a hierarchical neural architecture search method to find the optimal architecture from such a supernet.
As shown in Fig. \ref{supernet}, the predefined network comprises $M=9$ global nodes. Each node is a latent representation (e.g. a feature map in CNNs). Two nodes are connected by three directed edges (yellow, blue, and green lines) associated with some operations. Specifically, the edges with yellow line and blue line denote the none and skip-connect operators. The edge associated with the green line presents a cell where there are $N=4$ local nodes, each connected with 8 operators in Table \ref{table12}. The supernet thus comprises in total $M=9$ global nodes and  $N=4$ local nodes. 
In DARTS, we recall that a cell is a directed acyclic graph of a node, where each node represents a network
layer. In our supernet, the global nodes and local nodes cannot form a directed acyclic graph as the global nodes are not connected with local nodes. It is difficult, therefore, to directly optimize the supernet by DARTS. 

To ensure an effective search, the search is divided into two stages. In the first, the architecture of each local cell is determined by a fixed global architecture, while in the second, the global architecture of the supernet is optimized. In this way, the supernet architecture and local cells are alternatively learned to obtain the final architecture for the downstream task, as shown in Fig. \ref{fig2}. Similarly, the search spaces are split into two sub-spaces: global space $\mathcal{O}_G$ and local space $\mathcal{O}_L$, as shown in Table \ref{table12}. The former comprises three candidate operations, namely, learning cells, skip-connection and none, 
 {while the latter comprises 8 candidate operations, namely, skip-connection, none, 3$\times$1 max pooling ($max\_pool\_3$), 3$\times$1 average pooling($avg\_pool\_3$), 3$\times$1 and 5$\times$1 separable convolutions ($sep\_conv$), 3$\times$1 and 5$\times$1 dilated separable convolutions ($dil\_conv$). Among them, $dil\_conv$ comprises a ReLU activation layer, a depthwise convolution layer, a pointwise convolution layer, and a batch normalization layer. $sep\_conv$ consists of a ReLU activation layer, a depthwise convolution layer, and a batch normalization layer, repeated twice as a whole.} Note that the non-operation indicates a lack of connection between two nodes.
\vspace{-5pt}

\begin{table}[!htbp]
\centering
\caption{The Search Space of SuperNet}
\resizebox{0.9\linewidth}{!}{

\begin{tabular}{c| c c }
\toprule
\multicolumn{2}{ c }{\textbf{Total Search Space} $\mathcal{O}$} \\
\hline

\textbf{Global search space $\mathcal{O}_G$} & \textbf{Local search space $\mathcal{O}_L$}& \\
\hline

none & -  \\
\hdashline

skip-connect & -\\
\hdashline

\multirow{8}*{cell} & none   \\

 & skip-connect   \\

 & max\textunderscore pool\textunderscore 3\\
 
 & avg\textunderscore pool\textunderscore 3 \\

 & sep\textunderscore conv\textunderscore 3 \\

 & sep\textunderscore conv\textunderscore 5 \\
 
 & dil\textunderscore conv\textunderscore 3 \\

 & dil\textunderscore conv\textunderscore 5 \\

\bottomrule
\end{tabular}
}
\label{table12}
\vspace{-5pt}
\end{table}
To avoid overfitting and reduce the computation cost of the target network, we employ entropy\cite{Shannon1948entropy} as an evaluation metric to simplify the network structure obtained from the search stage. Algorithm. \ref{alg1} illustrates the overall search algorithm.

\subsection{Local Search}
As DARTS searches a cell and stacks multiple learning cells to form a network, all cells share the same architectural parameters. In our work, we predefine a supernet with $M$ global nodes, as shown in Fig. \ref{supernet}. The local cells are treated as directed edges in the supernet, so there are $M-1$ local cells, with each cell having two input nodes, an output node, and $N$ intermediate nodes.  {Each node represents a feature map, and each edge between nodes signifies an operation.} We fix the global architecture and optimize all local cells simultaneously. Let  $e^m_{i,j}$ be the edge from node $i$ to node $j$ in the $m$-$th$ cell, associated with input $x^m_i$ and its output presented as $e^m_{i,j}(x^m_i)$. The intermediate node $j$ is computed by Eq. \ref{eq1}:
\begin{align}
	 x^m_j=\sum_{i<j}e^m_{i,j}(x^m_i).
  \label{eq1}
\end{align}

Let $\mathcal{O}_L^m$ be a set of candidate operations (e.g., convolution, max pooling, and none) for the $m$-$th$ local cell with each operation denoting some function $e^m(\cdot)$. 
Based on DARTS\cite{Liu2018DARTSDA}, we relax the categorical selection of a specific operation to a softmax over all operations to obtain Eq. \ref{eq2}:
\begin{align}
	 \bar{e}^m_{(i,j)}(x) = \sum_{e^m \in \mathcal{O}_L^m} \frac{\exp(\alpha_{e^m_{(i,j)}})}{\sum_{e'^m \in \mathcal{O}_L^m} \exp(\alpha_{{e'^m}_{(i,j)}})} e^m(x).
	\label{eq2}
\end{align}
where each operation $e^m_{i,j}$ is related to a continuous coefficient $\alpha_{e^m_{(i,j)}}$.
 {Specifically, we transform the discrete action choices into a continuous probability distribution from 0 to 1 using the softmax function, which renders the entire search space differentiable.}
The architecture search task is then reduced to learning continuous variable $\alpha^*$ by minimizing $\mathcal{L}_{val}$ and the corresponding network weights $w^*$ by minimizing $\mathcal{L}_{train}$, with $\mathcal{L}_{train}$ and $\mathcal{L}_{val}$ the training and validation cross-entropy losses. The objective function for this task can be expressed as follows:
\begin{align}
	\min_{\alpha} \quad & \mathcal{L}_{val}(w^*(\alpha,\beta^*), \alpha) \label{eq3} \\
	\text{s.t.} \quad &w^*(\alpha,\beta^*) = \mathrm{argmin}_w \enskip \mathcal{L}_{train}(w, \alpha,\beta^*). \label{eq4}
\end{align}

However, it is expensive to evaluate the architecture performance after optimizing the parameters at each step. As for DARTS\cite{Liu2018DARTSDA}, gradient descent is used to update $\alpha$ by Eq. \ref{eq5}:

\begin{equation}
\begin{aligned}
&\nabla_\alpha \mathcal{L}_{val}(w^*(\alpha, \beta^*), \alpha) \\
&\approx \nabla_\alpha \mathcal{L}_{val}(w - \xi^\alpha \nabla_{w} \mathcal{L}_{train}(w, \alpha, \beta), \alpha),
\label{eq5}
\end{aligned}
\end{equation}
where $\xi^{\alpha}$ is the optimization learning rate and $w $ denotes the current weights of the search network.

Current search methods \cite{Liu2018DARTSDA}\cite{Chu2019FairDE}\cite{Chu2020DARTS-} stack cells with the same architecture parameters $\alpha$, which results in limited diversity, structure redundancy and  performance collapse\cite{Movahedi2022lamda}\cite{Zela2019UnderstandingAR}\cite{Chu2020DARTS-}. Our local search strategy does not share architectural parameters $\alpha$ between cells, as independent architectural parameters are learned for each cell, so as to improve the feature representation capacity of the target neural network.

\subsection{Global Search }

After searching the parameters of all local cells, the  global target network architecture becomes important to achieve good performance for the downstream task.  There are total $M$ nodes constituting a directed acyclic graph. The global space $\mathcal{O}_G$ includes three candidate operators, where the learning local cells are treated as operators. Similarly, the edge $E(i,j)$ represents the
information flow from node $i$ to $j$, which includes the candidate operations $r_{i,j}$ weighted by the architecture parameter $\beta_{i,j}$. The intermediate node gathers all inputs from the incoming edges and is obtained by Eq. \ref{eq6}:
\begin{align}
	 x_j=\sum_{i<j}r_{i,j}(x_i).
  \label{eq6}
\end{align}
To make the search space continuous,
the categorical choice of a particular operation is relaxed by a softmax over all operations $\mathcal{O}_G$ by Eq. \ref{eq7}:

\begin{align}
	 \bar{r}_{(i,j)}(x) = \sum_{r \in \mathcal{O}_G} \frac{\exp(\beta_{r_{(i,j)}})}{\sum_{r' \in \mathcal{O}_G} \exp(\beta_{{r'}_{(i,j)}})} r(x).
	\label{eq7}
\end{align}
The architecture optimization is converted to learning a set of continuous variables $\beta_{i,j}$, which can be solved with a triplet-level optimization:
\begin{align}
	\min_{\beta} \quad & \mathcal{L}_{val}(w^*(\alpha^*,\beta), \beta)  \\
	\text{s.t.} \quad &w^*(\alpha^*, \beta) = \mathrm{argmin}_w \enskip \mathcal{L}_{train}(w, \alpha^*,\beta). 
\end{align}

Similarly, $\beta$ is updated via gradient descent by Eq. \ref{eq10} :
\begin{equation}
\begin{aligned}
&\nabla_\beta \mathcal{L}_{val}(w^*(\alpha^*,\beta), \beta) \\
&\approx \nabla_\beta \mathcal{L}_{val}(w - \xi^\beta \nabla_{w} \mathcal{L}_{train}(w, \alpha,\beta), \beta). 
\label{eq10} 
\end{aligned}
\end{equation}

\subsection{Network Training}
After updating parameters $\alpha$ and $\beta$, the CNN with weights $w$ associated with the resulting architecture is trained by minimizing the training set loss.
\begin{align}
	w^*(\alpha^*, \beta^*) = \mathrm{argmin}_w \enskip \mathcal{L}_{train}(w, \alpha^*,\beta^*). 
\end{align}
Specifically, the weights $w$ are updated by Eq. \ref{eq12}:
\begin{equation}
\nabla_w \mathcal{L}_{train}(w - \xi^w \nabla_{w} \mathcal{L}_{train}(w, \alpha^*,\beta^*), w)\ 
\label{eq12}.
\end{equation}

\subsection{Alternative Search}
The global search and local search are optimized greedily to obtain the optimal architecture. Although the global architecture may be optimal w.r.t the local architecture,  the local cell is not optimal w.r.t the global supernet. To overcome this issue, we fix the global hyperparameters $\beta$ to achieve
the local search again, with the target network architecture obtained by alternatively performing this process for $K$ times, as shown in Fig. \ref{fig2}.

After obtaining the optimal architecture parameters $\beta^*$  and $\alpha^*$, the target discrete architecture $\phi$ is derived by replacing the $\bar{e}^m_{(i,j)}(x)$ with $e^m_{i,j}=argmax_{e^m\in \mathcal{O}_L^m}\alpha^m_{i,j}$ ($m=1,2,...,M-1$) as well as replacing the  $\bar{r}_{(i,j)}(x)$ with $r_{i,j}=argmax_{r\in \mathcal{O}_G}\beta_{i,j}$.


\subsection{Transfer Entropy}
Traditional  DARTS-based methods \cite{Liu2018DARTSDA}\cite{Chu2019FairDE}\cite{Movahedi2022lamda}\cite{Chu2020DARTS-} manually stack multiple cells to form a network for the downstream task.  For eye movement recognition, a deep neural network with deeper layers is prone to overfitting as the data are relatively scarce. To reduce the redundancy of the target network,  \textit{transfer entropy} is proposed to compute the information amount in each layer and remove the layers with relatively low entropy.

\begin{figure*}[htbp]
    \centerline{\includegraphics[scale=0.9]{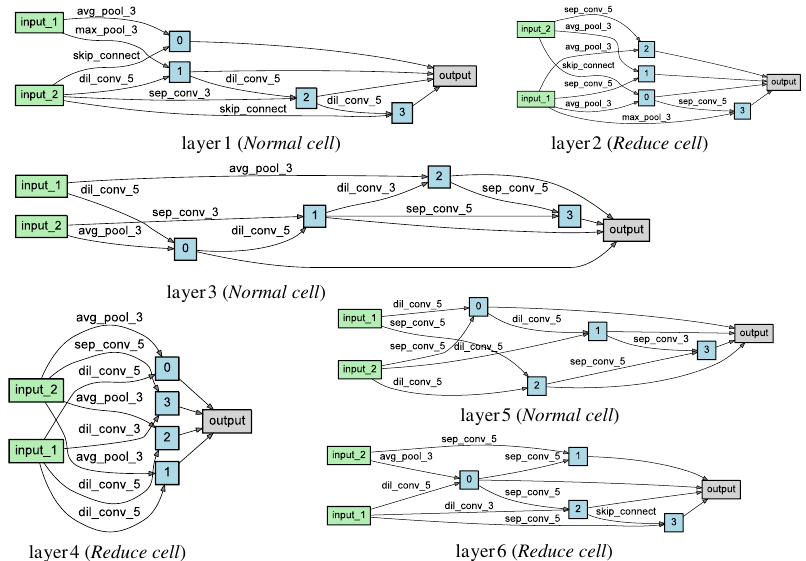}}
    \caption{ {Visualization of network structure obtained based on JuDo1000 database.}}
    \label{fig7}
    \vspace{-10pt}
\end{figure*}

Entropy can be used to quantify the information content of a network\cite{Guan2019entropy-new}.  {In the context of neural networks, TE can be employed to evaluate the information flow between layers or neurons, helping identify redundant connections that contribute minimally to the overall network performance.} For a given probability distribution $p(z), z \in Z$, the output entropy  of a given layer can be calculated as:
\begin{equation}
H(Z) = - \int p(z)\log p(z)df, z \in Z.
\end{equation}
Following \cite{Lin2024MLPCB}, the entropy calculation is simplified by employing a Gaussian distribution as the probability distribution of the intermediate features in the layer, the layer entropy is then approximated by Eq. \ref{eq14}.
\begin{equation}
\begin{aligned}
H(Z) = - \mathbb{E} \ [\log \mathcal{N}(\mu,\sigma^2)]  \\
 = - \mathbb{E} \ [\log [\frac{1}{2 \pi \sigma^2}e^{-\frac{1}{2\sigma^2}(z-\mu)^2}]] \\
 = \log(\sigma)+\frac{1}{2}\log(2\pi)+\frac{1}{2},
 \label{eq14}
\end{aligned}
\end{equation}
where $Z$ is the feature set and $Z\sim \mathcal{N} (\mu, \sigma^2)$, with $\sigma$ and $\mu$ the standard deviation and mean of the feature set $z\in Z$. From Eq. \ref{eq14}, $H(Z)$ is proportional to $log(\sigma)$ with two additional constants. Without loss of generality, we do not consider the two constants in the application. Specifically, a batch of data is input to network model $\phi$ to obtain feature set $Z$. Assume that $Z_i$ includes all features from the $ith$ layer of $\phi$ and $Z_{i,j}$ are the feature maps of the $jth$ channel in the $ith$ layer. We compute the sum of the logarithm of the standard  deviation of each feature channel by Eq. \ref{eq15}:

\begin{equation}
\begin{aligned}
 H_{\sigma}=\sum_j{log[\psi(Z_{i,j})]},
\end{aligned}
\label{eq15}
\end{equation}
where $\psi(Z_{i,j})$ is the
standard deviation of the $jth$ channel of the feature set $Z_i$ of the $ith$ layer. 
As the entropy $H(Z) $ is proportional to  $log(\sigma)$,  it is also proportional to the sum of $log(\sigma)$, namely $H_{\sigma}(Z)$. In other words, $H_{\sigma}(Z)$ can be treated as the entropy $H(Z)$ of a given layer. The transfer entropy aims to measure the information amount of directed transfer between two layers \cite{Lin2024MLPCB}. Concretely, the transfer entropy (TE) of $ith$ layer in network $\phi$ can be computed based on output maps of  the $ith$ layer and $i-1th$ layer by Eq. \ref{eq16}:
\begin{equation}
\begin{aligned}
{TE}_i = |\ H(Z_i) - H(Z_{i-1}) \ | .\\
\end{aligned}
\label{eq16}
\end{equation}
In this way, we obtain the transfer entropy of all layers in network $\phi$. The layers
with minimum transfer entropy are then removed from network $\phi$ and we compute the accuracy of the resulting network model on the validation set. At each round, we traverse all the layers and remove the layer with a minimum transfer entropy by greedy search. The procedure is repeated until the accuracy of the resulting model starts to reduce, at which stage we get the final network $\phi^\mathcal{*}$.  {Fig. \ref{fig7} shows the details of each layer in the search network. As shown in Fig. \ref{fig7}, the searching network consists of six layers (cells), e.g. three normal cells and three reduction cells. From Fig. \ref{fig7}, we can see that all cells have different architecture parameters, which demonstrates that our approaches can improve the diversity of searching network.  }

\section{Experiments}
To evaluate our approach, we performed extensive experiments on three public datasets: GazeBase \cite{Griffith2021GazeBaseAL}, JuDo1000\cite{Makowski2020BiometricIA}, and EMglasses\cite{Qin2024EmMixformerMT}. We compare EM-DARTS with state-of-the-art approaches, including five hand-designed DL methods for eye movement recognition, namely DEL \cite{Jger2019DeepEB}, Dense LSTM\cite{Taha2023EyeDriveAD}, EKYT\cite{Lohr2022EyeKY}, RP-Inception\cite{RP-InceptionV3} and EmMixformer\cite{Qin2024EmMixformerMT}. Also, four DARTS-based methods, namely the classical DARTS\cite{Liu2018DARTSDA} algorithm as well as its improvement version, Fair-DARTS\cite{Chu2019FairDE}, DARTS-\cite{Chu2020DARTS-} and $\Lambda$-DARTS\cite{Movahedi2022lamda} are reproduced for comparison.
 {All experiments were conducted on a high-performance server running Ubuntu 18.04, equipped with an Intel(R) Xeon(R) Platinum 8358 CPU, 975 GB RAM, and an NVIDIA A100 Tensor Core GPU. The code was implemented in Python 3.9.12 using the PyTorch 1.12.1 framework with CUDA 12.2 support.}


\subsection{Datasets}
\paragraph{GazeBase}
The GazeBase dataset \cite{Griffith2021GazeBaseAL} comprises eye movement data from 322 volunteers using the Eye-Link1000 eye-tracker. Each volunteer in participated seven tasks with different stimulus materials, namely a Randomized scanning task (RAN), a Reading task (TEX), two Video viewing tasks (VD1 and VD2), a Gaze task (FXS), a Horizontal scanning task (HSS), and an Eye-driven game task (BLG).  In our experiments, we use four sub-datasets from GazeBase, namely RAN, TEX, FXS, and HSS, for recognition purposes.  {Due to its extensive participant pool and diverse task settings, GazeBase provides a robust foundation for hypothesis validation in eye movement biometrics and advanced machine learning analysis of gaze signals. Its varied experimental tasks allow for comprehensive testing of models across a wide range of scenarios, enhancing its applicability to both theoretical research and practical implementations.} 
\paragraph{JuDo1000}
The JuDo1000 dataset\cite{Makowski2020BiometricIA} comprises eye movement data from 150 volunteers who participated in four experiments, separated by a time interval of at least one week. The data were recorded by an EyeLink Portable Duo eye-tracker with a sampling frequency of 1000 Hz. During the experiments, a black dot appeared on the screen at five random locations, with duration varying between 250, 500, and 1000 ms. Participants' heads were immobilized by chin and forehead mounts.  {JuDo1000, with its controlled acquisition conditions and precise measurements, has established itself as a benchmark dataset with minimal noise. This precision makes it an ideal foundation for developing accurate mathematical models of eye movement and for training large-scale, high-performance recognition systems.}
\paragraph{EMglasses}
 {While high-frequency, low-noise data can achieve excellent recognition results, acquiring such data is challenging in practical applications.} 
As signals with sampling frequencies above 30 Hz were shown to meet the requirements for eye movement recognition \cite {Holland2012BiometricVV },  we built the EMglasses dataset \cite{Qin2024EmMixformerMT} comprising 203 volunteers, collected using a Tobii Pro Glasses3 eye-tracker with a low-frequency eye-tracking device (50Hz).  {The device captures subtle head movements during data collection. As a result, the dataset introduces greater variability and complexity, making it more challenging for eye movement recognition while providing a realistic representation of practical scenarios.} In our experiments, a random dot of 1-second duration was used as a stimulus to participants who observed twice at different times. 

\subsection{Data Processing}
As there are significant differences in velocity between saccadic and fixational eye movements,  a global normalization would cause the slow fixational
drift and tremor to near zero, which results in missing high-frequency information for slow eye movements. As in \cite{Qin2024EmMixformerMT}\cite{Makowski2021DeepEyedentificationLiveOB}, we split the eye movement data into fast and slow components. Fast data are specifically related to rapid eye movements, while slow data focus on the subtle movements associated with fixational drift and tremor. 

First, the eye gaze point coordinates data X = ($x_{1},..., x_{t}$) and Y = ($y_{1},..., y_{t}$) are transformed into velocity data $\theta$ at each sampling point. The  velocity data $\theta_s^{(i)}$ of the $i$th point are computed using Eq. \ref{eq17}:
\begin{equation}
\theta^{(i)}_s = \frac{|s^{(i)}-s^{(i-1)|}}{t^{(i)}-t^{(i-1)}},
\label{eq17}
\end{equation}
where $s^{(i)}$ represents the $i$ th point coordinates  $x_i$ or $y_i$, and the $\theta_{x}$ and $\theta_{y}$ denote the horizontal velocity sequence and vertical velocity sequence, respectively. Then, the $NaN$ value is replaced with 0 in $\theta_{x}$ and $\theta_{y}$. 
Furthermore, the velocity data $\theta_s$ are divided into two parts: fast data $\Phi_{fast}$ and slow data $\Phi_{slow}$, using the velocity threshold $v_{min}$. 
For the fast data $\Phi_{fast}$, velocities below the minimum threshold $v_{min}$ are truncated and $Z-Score$ is applied for normalization:
\begin{equation}
\Phi_{fast}^{(i)}{(\theta^i_x, \theta^i_y)}=\left\{\begin{matrix}
(Z(\theta^i_x),Z(\theta^i_y))\quad\quad\quad\quad otherwise\\

\quad Z(0) \quad\quad if\sqrt{(\theta^i_x)^{2}+(\theta^i_y)^{2}}< v_{min}
\end{matrix}\right.,
\end{equation}

\begin{equation}
Z_{i}=(\theta^{i}-\mu)/\sigma_i,
\end{equation}
where $\theta^{i}$ is the velocity at the $i$-th sampling point, and $\mu$ and $\sigma_i$ are the mean and variance of the sequence $\theta$, respectively.

For the slow data $\Phi_{slow}$, we use the hyperbolic tangent function to normalize the $\theta_x$ and $\theta_y$ to a range of $(-1, +1)$:

\begin{equation}
\Phi_{slow}^{(i)}{(\theta^i_x, \theta^i_y)} = (tanh(c\theta^i_x),tanh(c\theta^i_y)).
\end{equation}
The velocity threshold $v_{min} $ and the scaling factor $c$ are fixed at 40$^\circ$/s and 0.02. Finally, the $\Phi_{fast}$ and $\Phi_{slow}$ are concatenated in the channel dimension before being forwarded to the network. 

\RestyleAlgo{ruled}

\begin{algorithm*}
\caption{EM-DARTS}\label{alg1}
\KwIn{The $train$ and $val$ data; Architecture weights $\alpha$; Network weights $\omega$; Input weight $\beta$; Search epochs $K$;}
\KwOut{Final network $\phi^\mathcal{*}$;}

Initialize $\alpha$ and $\beta$\ and determine the search space;

    
Define a supernet and initialize network weights $\omega$\;

\For{k $\in$ [1, K] }
    {
         Sample $batch$ $\in$ $val$\;
         \quad\quad$Cell_i$ independently updates the $\alpha_i$ by $\nabla_\alpha\mathcal{L}_{val}$\;
         \quad\quad Update the $\beta_i$ by $\nabla_\beta\mathcal{L}_{val}$\;
         Sample $batch$ $\in$ $train$\;
         \quad\quad Update $\omega$ by  $\nabla_\omega\mathcal{L}_{train}$\;
        obtain the model architecture based on the learned $\alpha^*$ and $\beta^*$\;
         
    }
Output the target networks $\phi$ \;
\While{The accuracy of simplifying network  on $val$ data increases }{
        Compute the transfer entropy of all layers in the network $\phi$\;
        Remove the layer with a minimum transfer entropy from $\phi$;\\
        Train the resulting network $\phi'$\ and compute the accuracy on $val$ data;\\
        Compute the difference of networks before and after simplifying;
    }
Output the final networks $\phi^\mathcal{*}$\;
\end{algorithm*}

\subsection{Experimental Settings}
 
To assess EM-DARTS, we evaluated the performance of our approach on three datasets, namely GazeBase \cite{Griffith2021GazeBaseAL}, JuDo1000\cite{Makowski2020BiometricIA}, and EMglasses\cite{Qin2024EmMixformerMT}, with each divided into training set and test set according to the collecting time sessions.

GazeBase has four subsets, named RAN, TEX, FXS, and HSS, each designed for recognition tasks and comprising nine rounds (Rounds 1-9) of data collected over 37 months, with two collections per round. During the data collection process, only a portion of the participants from the last round provided data for the subsequent round.  {We focus on the first round of data, which includes the largest number of participants, comprising 644 sequences from 322 participants, and we divide the two acquisition phases into training (322 $\times$ 1) and test sets (322 $\times$ 1). For each subset, we select data from the first session for training and the second session for testing, which results in four training and test sets for the first round of RAN, TEX, FXS, and HSS.}


JuDo1000 is collected from 150 participants, each providing data four times. The data from the first three times are used for training and the data from the fourth for test. Accordingly, the training set comprises 450 sequences (150 $\times$ 3), and the test set 150 sequences (150 $\times$ 1).

For EMglasses, 203 participants provided eye movement data collected from two different time sessions (203 $\times$ 2). The data from the first and second sessions were selected as training and test sets, respectively, with each comprising 203 sequences (203 $\times$ 1).

 {The data in the three datasets were collected for several seconds, which resulted in long sequences. To facilitate training, we divide each sequence into sub-sequences according to the sampling frequency, each sub-sequence is one second of data. The sampling frequency of GazeBase is 1000 Hz. In the RAN subset, the training and test sets were split into 31626 and 31511 samples, respectively. In the TEX subset, the training and test sets contain 18366 and 18305 samples, respectively. In the FXS subset, the training and test sets contain 4519 and 4532 samples, respectively. In the HSS subset, the training and test sets contain 31709 and 31755 samples, respectively. The sampling frequency of JuDo1000 is 1000 Hz and the training and test sets contain 141837 and 47221 samples, respectively and the sampling frequency of EMglasses is 50 Hz and the training and test sets contain 2396 and 2407 samples, respectively.} 
The eye movement recognition approaches,  DEL \cite{Jger2019DeepEB}, Dense LSTM\cite{Taha2023EyeDriveAD}, EKYT\cite{Lohr2022EyeKY} and RP-Inception\cite{RP-InceptionV3}, are trained on the training set and evaluated on the test set is reported for comparison. To train the architecture search approaches, DARTS\cite{Liu2018DARTSDA}, Fair-DARTS\cite{Chu2019FairDE}, DARTS-\cite{Chu2020DARTS-} and $\Lambda$-DARTS\cite{Movahedi2022lamda}, the data of each subject in the training set is further split in two parts, by selecting 70\% sequences for training and 30\%  for validation. As a result, there are three subsets: new training set, validation set, and test set for GazeBase \cite{Griffith2021GazeBaseAL}, JuDo1000\cite{Makowski2020BiometricIA}, and EMglasses\cite{Qin2024EmMixformerMT}. The new training set and validation set are used for network architecture search. After obtaining the optimal architecture parameters, the resulting network is re-trained on the original training set (new training set + validation set) and evaluated on the test set.  {Details of training and testing data sets are presented by Table \ref{table_dataset}.}

\begin{table}[!htbp]
\centering
\caption{The amount of data points in each set}
\resizebox{0.9\linewidth}{!}{

\begin{tabular}{c | c c | c c}
\toprule
& \multicolumn{2}{ c }{\textbf{Search Phase}} & \multicolumn{2}{ c }{\textbf{Train Phase}} \\
\hline

\textbf{Dataset} & New training set & Validation set & Training set & Test set \\
\hline

JuDo1000 & 99286 & 42551 & 141837 & 47221 \\

RAN & 22138 & 9488 & 31626 & 31511\\

TEX & 12835 & 5501 & 18336 & 18305\\

FXS & 3163 & 1356 & 4519 & 4532   \\

HSS & 22196 & 9513 & 31709 & 31755\\
 
EMglasses & 1677 & 719 & 2396 & 2407\\

\bottomrule
\end{tabular}
}
\label{table_dataset}
\end{table}

For a fair comparison, all the methods follow the same network parameter settings. For eye movement recognition methods\cite{Makowski2021DeepEyedentificationLiveOB,Lohr2022EyeKY,Taha2023EyeDriveAD,RP-InceptionV3,Qin2024EmMixformerMT}, the initial learning rate is set to 0.0002 and the batch size is fixed at 64. The maximum number of training epochs is set to 1000. Early stopping is configured with a patience of 30 epochs, and the weight decay is set to 1e-5.
For benchmarking NAS methods\cite{Liu2018DARTSDA,Chu2019FairDE,Chu2020DARTS-,Movahedi2022lamda} and our approach, all the methods perform 50 search epochs and 300 training epochs. 
In the search phase, the batch size is set to 32 for training and 128 for validation. The learning rate of weight $w$ decays from 0.025 to 0.001 with cosine annealing, and SGD is selected as the optimizer with a momentum of 0.9 and a weight decay of $5 \times
10^{-4}$. For updating of $\alpha$ and $\beta$, the learning rates  are set to $3 \times 10^{-4}$, Adam as the  optimizer with a weight decay of $1 \times 10^{-3}$, and the drop-path
\cite{Larsson2016FractalNetUN} rate is set to 0.3. 

To assess our approach, we report the Equal Error Rate (EER) mean at the last 10 epochs on the test set. EER is defined as the error rate at which the False Acceptance Rate (FAR) equals the False Rejection Rate (FRR). 
 {
The True Positive Rate (TPR) is computed by $1-$FRR. TPR (Recall rate) represents the proportion of correctly identified positive classes' overall actual positive instances, while the FPR represents the proportion of negative classes incorrectly identified as positive out of all actual negative classes. We utilize Receiver Operating Characteristic (ROC) curves to illustrate the relationship between the true positive rate and the false positive rate across various thresholds, thereby demonstrating how the model's performance varies with different threshold values. Furthermore, to provide a comprehensive evaluation of the model's performance, we include the Precision \textit{vs.} Recall (PR) curve, which depicts Precision and Recall at different thresholds.} 
Lower EER values indicate better verification performance. Additionally, we report the FRR at different FARs ($10^{-1}$, $10^{-2}$, and $10^{-3}$) for a more comprehensive model evaluation. 
\begin{figure}[htbp]
\centerline{\includegraphics[scale=0.55]{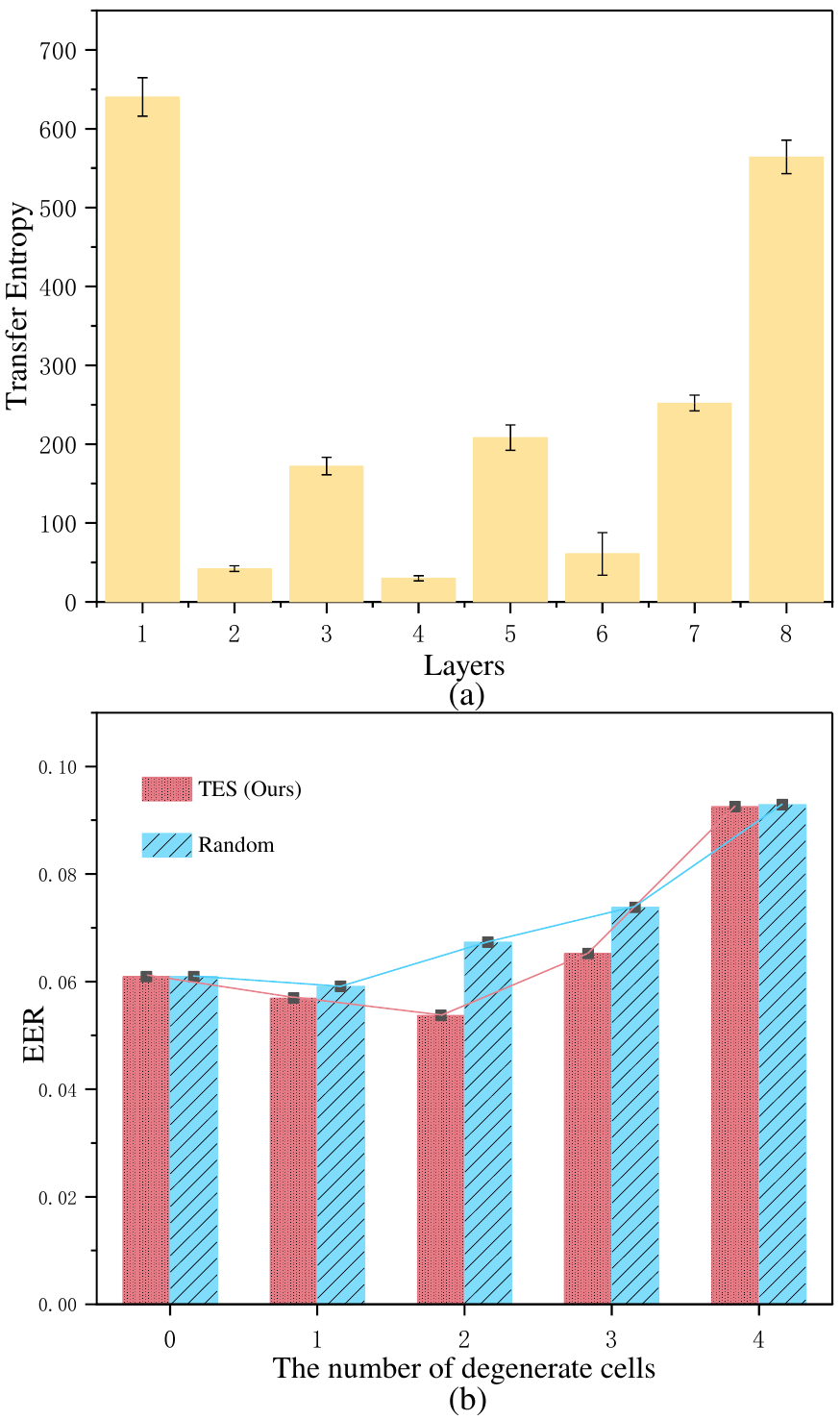}}
    \caption{ {The experimental results on the RAN dataset. (a) The transfer entropy of each layer. (b) The EER of resulting models by removing the layers with the random degradation strategy and transfer entropy.}}
    \label{fig8}
    \vspace{-10pt}
\end{figure}

\begin{figure*}[htbp]
\centerline{\includegraphics[scale=0.83]{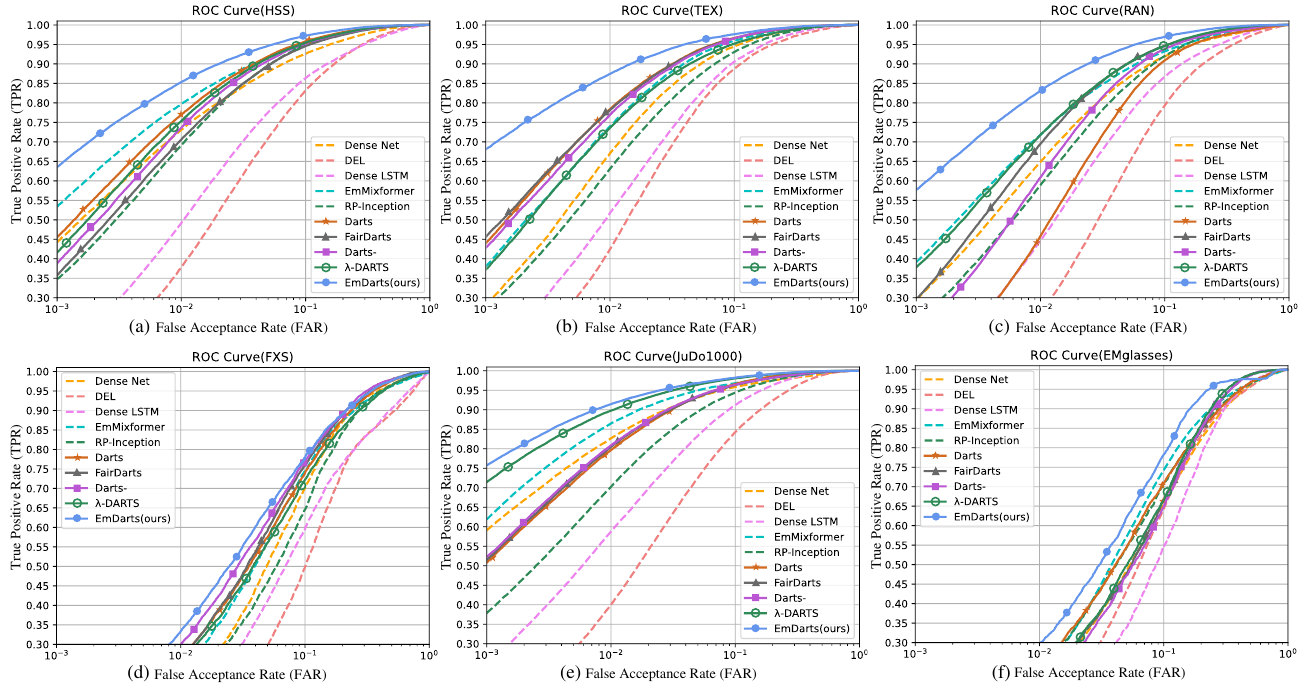}}
    \caption{ROC curve of various approaches on (a) HSS dataset, (b) TEX dataset, (c) RAN dataset, (d) FXS  dataset,  (e) JuDo1000 and  {(f) EMglasses dataset}.}
    \label{fig5}
    \vspace{-5pt}
\end{figure*}

\begin{figure*}[htbp]
\centerline{\includegraphics[scale=0.75]{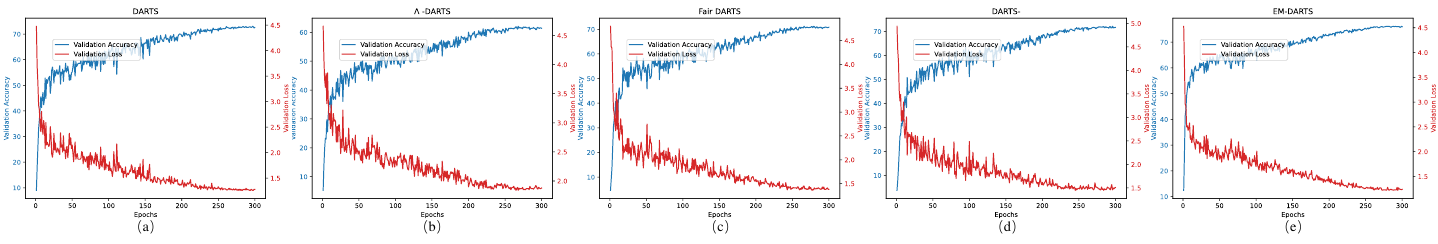}}
    \caption{ {Curve of validation accuracy and loss over epochs on TEX dataset. (a) DARTS, (b) $\Lambda$-DARTS, (c) Fair DARTS, (d) DARTS-, (e) EM-DARTS.}}
    \label{fig9}
    \vspace{-10pt}
\end{figure*}

\subsection{ Verification Performance of Transfer Entropy Strategy.}
Generally, a CNN with more layers is prone to overfitting, while fewer layers may result in underfitting. This experiment focuses visually on assessing the efficacy of the Transfer Entropy Strategy (TES) in simplifying the network structure.  The experiments are conducted on the RAN sub-dataset in GazeBase and we report the results of CNN models after their simplification by our TES approach and random selection scheme.  {First, the optimal CNN architecture is obtained by our hierarchical neural architecture search, and we computed the TE of each layer based on the feature maps during training. Subsequently, we sort the transfer entropy values of all layers and remove the layer with the lowest transfer entropy by applying a `skip-connection'. Instead of using a specific threshold, which would necessitate additional restrictions and evaluations of the range of transfer entropy.} 
 {The resulting architecture is trained and the EER on the test set is reported for comparison. The results of the two approaches are illustrated in Fig. \ref{fig8}. Fig. \ref{fig8}(a) illustrates the transfer entropy across all layers in the search model, while Fig. \ref{fig8}(b) shows the EER of the resulting model after removing varying numbers of layers using both the TE strategy and the random cropping strategy, respectively.}
We observe that the performance of the resulting network is optimal after removing two layers from the search CNN model by our TES. When filtering three layers with our TES, the verification error of the resulting model is increased. This is because such a model with fewer layers may suffer from underfitting. However, when a random amount of layers are removed from the search model, the performance of the resulting model is further degraded. Therefore, the results in Fig. \ref{fig8} show that our TES, by effectively computing the information amount of each layer and filtering them accordingly, achieves better performance than random selection schemes.  {In addition, in the ablation experiment presented in Section IV, we compare the computational overhead of the search process across different layers. As shown in Table \ref{table14}, our strategy of employing TE for pruning not only reduces computational overhead but also enhances network performance.}

\begin{figure}[htbp]
\centerline{\includegraphics[scale=0.82]{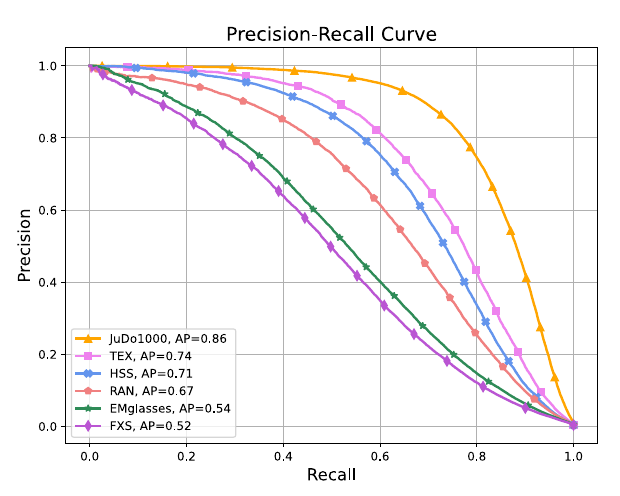}}
    \caption{ {Precision vs. recall curves of EM-DARTS.}}
    \label{PR}
    \vspace{-15pt}
\end{figure}
\subsection{Verification Performance of EM-DARTS}
This section assesses our EM-DARTS on the three eye movement datasets. We reproduce state-of-the-art eye movement recognition algorithms, namely DEL\cite{Makowski2021DeepEyedentificationLiveOB},
EKTY\cite{Lohr2022EyeKY}, Dense LSTM\cite{Taha2023EyeDriveAD}, RP-Inception\cite{RP-InceptionV3}, EmMixformer\cite{Qin2024EmMixformerMT} for comparison. Also, we implement NAS algorithms, DARTS \cite{Liu2018DARTSDA}, Fair DARTS\cite{Chu2019FairDE}, DARTS-\cite{Chu2020DARTS-} and $\Lambda$-DARTS\cite{Movahedi2022lamda}, to benchmark the performance of our approach. Tables \ref{table2}-\ref{table7} present the EER and FRR at different FARs of the various approaches. Fig. \ref{fig5} illustrates the corresponding ROC curves  {Fig. \ref{fig9} displays the plots of Validation Loss and Validation Accuracy on the validation set versus training epochs and Fig. \ref{PR} demonstrates Precision and Recall of EM-DARTS at different thresholds on different datasets, with AP denoting the average precision.
To illustrate the layer details of the network, Fig. \ref{fig7} visualizes the searched structures on JuDo1000.}

From the results in Tables \ref{table2}-\ref{table7}, we observe that our EM-DARTS outperforms current approaches and achieves the lowest verification error on the four subsets of GazeBase, JuDo1000, and EMglasses. Also, we observe that our approach achieves the highest recognition accuracy at different FRRs from Fig. \ref{fig5}.  {In addition, the experimental results in Fig. \ref{fig9} demonstrate that the network searched by our method achieves stable convergence.}
 
Overall, the NAS-based approach achieves lower verification errors than the manually designed CNN on the three datasets. This can be explained by the following facts. A manually designed CNN requires rich prior knowledge. However, human experts cannot access all candidate architectures in the CNN search space due to their limited knowledge and experience, especially for huge architectural spaces. In addition, a manually designed CNN lacks generalization on different tasks.  For example, a CNN manually designed for specific tasks may show poor performance for other tasks or datasets. An automatically designed CNN, by contrast, explores a huge space of candidate deep neural network architectures by minimizing the verification error on the evaluation set. The NAS-based automatically designed architecture avoids the need for manual architecture design, with the risk of discarding optimal networks for classification. Instead, an optimal architecture for different tasks is produced, which improves the capacity of generalization.  This explains why NAS-based CNNs achieve higher recognition accuracy on the three datasets. Compared to NAS-based CNNs, our approach achieves the lowest verification accuracy, which can be attributed to the following facts. State-of-the-art NAS-based approaches, namely DARTS \cite{Liu2018DARTSDA}, Fair DARTS\cite{Chu2019FairDE}, DARTS-\cite{Chu2020DARTS-} and $\Lambda$-DARTS\cite{Movahedi2022lamda}, search an optimal cell architecture based on a shallow network and stack multiple cells together to form a deeper network as the target network for downstream tasks, which raises two issues. First, all cells for stacking share the same architecture parameters, which reduces the representation capacity of the target network. Second, the search for optimal operations may work well in a shallow architecture but can show poor performance on the evaluation stage. Our approach, by contrast, directly defines a supernet and searches the optimal architecture for each cell, which increases the diversity of the network. At the same time, the search and evaluation stages adopt the same architecture, which avoids the gap between the architecture depths in the search and evaluation scenarios. Additionally, we compute the transfer entropy to reduce the redundancy of the searched CNN, to further improve the performance of the target network. 
 
Traditional DARTS searches the reduction cell and normal cell by minimizing the loss of the shallow CNN network, where the cells share the same architecture weights $\alpha$.  In general, there are differences between the cells in deeper layers and in shallow layers. Our approach, by contrast, can search the cells with different architectures. Similarly to traditional  DARTS, our approach can be extended to large deep neural network searches by stacking such different cells to form a deeper CNN model. The resulting model acquires a more robust representation capacity than when stacked by the same cell.   \vspace{-5pt}

\begin{table}[!htbp]
\caption{Results of various approaches on the HSS database}
\centering
\resizebox{\columnwidth}{!}{
\begin{tabular}{ p{0.9cm} c c c c c }
\toprule
\multirow{2}*{Approach} & \multirow{2}*{HSS} &\multirow{2}*{EER} & \multicolumn{3}{c}{FRR@FAR}\\
\cline{4-6}
& &  & $10^{-1}$ & $10^{-2}$  & $10^{-3}$\\
\hline
\multirow{5}*{DL-based } & DEL\cite{Makowski2021DeepEyedentificationLiveOB} &  0.1309 & 0.1680 & 0.6217 & 0.9087\\

& Dense LSTM\cite{Taha2023EyeDriveAD} & 0.1191 &0.1365 & 0.5082 & 0.8382\\

& RP-Inception\cite{RP-InceptionV3}&0.0727&0.0514&0.3085&0.6521\\

& EKYT\cite{Lohr2022EyeKY}& 0.0839 & 0.0739 & 0.2691 & 0.5575\\

& EmMixformer\cite{Qin2024EmMixformerMT} & 0.0673 &0.0502 & 0.2032 & 0.4659\\
\hline 
\multirow{5}*{NAS-based}& DARTS \cite{Liu2018DARTSDA} & 0.0642 & 0.0423 & 0.2307 & 0.5448\\

& Fair DARTS\cite{Chu2019FairDE} & 0.0736 & 0.0544 & 0.2939 & 0.6424\\

& DARTS-\cite{Chu2020DARTS-} & 0.0686 & 0.0482 & 0.2634 & 0.6119\\

& $\Lambda$-DARTS\cite{Movahedi2022lamda} & 0.0667 & 0.0456 & 0.2465 & 0.5839 \\

& \textbf{EM-DARTS (Ours)}  & \textbf{0.0520} & \textbf{0.0278} & \textbf{0.1464} & \textbf{0.3645}\\

\bottomrule
\end{tabular}
}
\vspace{-12pt}
\label{table2}
\end{table}
\begin{table}[!htbp]
\caption{Results of various approaches on the TEX database}
\centering
\resizebox{\columnwidth}{!}{
\begin{tabular}{ p{0.9cm} c c c c c }
\toprule
\multirow{2}*{Approach} & \multirow{2}*{TEX} &\multirow{2}*{EER} & \multicolumn{3}{c}{FRR@FAR}\\
\cline{4-6}
& &  & $10^{-1}$ & $10^{-2}$  & $10^{-3}$\\
\hline
\multirow{5}*{DL-based } & DEL\cite{Makowski2021DeepEyedentificationLiveOB} & 0.1060 & 0.1128 & 0.5750 & 0.9141\\

& Dense LSTM\cite{Taha2023EyeDriveAD}& 0.0971 & 0.0945 & 0.4824 & 0.8456 \\

& RP-Inception\cite{RP-InceptionV3} & 0.0837 & 0.0695 & 0.3688 & 0.7311  \\

& EKYT\cite{Lohr2022EyeKY}& 0.0736 & 0.0551 & 0.3293  & 0.7175 \\

& EmMixformer\cite{Qin2024EmMixformerMT} & 0.0635 & 0.0407 & 0.2603 & 0.6193\\

\hline 

\multirow{5}*{NAS-based} & DARTS \cite{Liu2018DARTSDA} & 0.0580 & 0.0338 & 0.2160 & 0.5606\\

& Fair DARTS\cite{Chu2019FairDE} & 0.0583 & 0.0347 & 0.2183 & 0.5445 \\

& DARTS-\cite{Chu2020DARTS-} & 0.0596 & 0.0343 & 0.2304 & 0.5717 \\

& $\Lambda$-DARTS\cite{Movahedi2022lamda} & 0.0688  & 0.0481  & 0.2641  & 0.6277 \\

& \textbf{EM-DARTS (Ours)} & \textbf{0.0453} & \textbf{0.0240} & \textbf{0.1254} & \textbf{0.3193}\\

\bottomrule
\end{tabular}
}
\vspace{-12pt}
\label{table3}
\end{table}
\begin{table}[!htbp]
\caption{Results of various approaches on the RAN database}
\centering
\resizebox{\columnwidth}{!}{
\begin{tabular}{ p{0.9cm} c c c c c }
\toprule
\multirow{2}*{Approach} & \multirow{2}*{RAN} &\multirow{2}*{EER} & \multicolumn{3}{c}{FRR@FAR}\\
\cline{4-6}
& &  & $10^{-1}$ & $10^{-2}$  & $10^{-3}$\\
\hline
\multirow{5}*{DL-based } & DEL\cite{Makowski2021DeepEyedentificationLiveOB} & 0.1436 & 0.2066 & 0.7383 & 0.9645\\

& Dense LSTM\cite{Taha2023EyeDriveAD} & 0.1161 & 0.1329 & 0.5529 & 0.8846\\

& RP-Inception\cite{RP-InceptionV3} & 0.0893 & 0.0800& 0.4108 & 0.7591 \\

& EKYT\cite{Lohr2022EyeKY}& 0.0885 & 0.0807 & 0.3513 & 0.7045\\

& EmMixformer\cite{Qin2024EmMixformerMT} & 0.0801 & 0.0680 & 0.2818 & 0.2818\\

\hline 

\multirow{5}*{NAS-based} & DARTS \cite{Liu2018DARTSDA} & 0.0956 & 0.0910 & 0.5412 & 0.8987 \\

& Fair DARTS\cite{Chu2019FairDE} & 0.0720  & 0.0532  & 0.3048 & 0.7038 \\

& DARTS-\cite{Chu2020DARTS-} & 0.0788 & 0.0617 & 0.3907 & 0.7915  \\

& $\Lambda$-DARTS\cite{Movahedi2022lamda} & 0.0726  & 0.0540  &  0.2816 & 0.6221 \\

& \textbf{EM-DARTS (Ours)}  & \textbf{0.0537} & \textbf{0.0304} & \textbf{0.1686} & \textbf{0.4226} \\

\bottomrule
\end{tabular}
}
\vspace{-12pt}
\label{table4}
\end{table}
\begin{table}[!htbp]
\caption{Results of various approaches on the FXS database}
\centering
\resizebox{\columnwidth}{!}{
\begin{tabular}{ p{0.9cm} c c c c c }
\toprule
\multirow{2}*{Approach} & \multirow{2}*{FXS} &\multirow{2}*{EER} & \multicolumn{3}{c}{FRR@FAR}\\
\cline{4-6}
& &  & $10^{-1}$ & $10^{-2}$  & $10^{-3}$\\
\hline
\multirow{5}*{DL-based } & DEL\cite{Makowski2021DeepEyedentificationLiveOB} & 0.2229 & 0.5022  &  0.9358 & 0.9936\\

& Dense LSTM\cite{Taha2023EyeDriveAD} & 0.2191 & 0.4047 &0.8831  &0.9832 \\

& RP-Inception\cite{RP-InceptionV3} & 0.1809 & 0.3546 & 0.8259 & 0.9576\\

& EKYT\cite{Lohr2022EyeKY}& 0.1666 & 0.2979 & 0.8323  & 0.9821 \\

& EmMixformer\cite{Qin2024EmMixformerMT} & 0.1578 & 0.2549 & 0.7654 & 0.9539 \\

\hline 

\multirow{5}*{NAS-based} & DARTS \cite{Liu2018DARTSDA} & 0.1609 &0.2606  & 0.7370 & 0.9402\\

& Fair DARTS\cite{Chu2019FairDE} & 0.1547  & 0.2421  & 0.7330 & 0.9481 \\

& DARTS-\cite{Chu2020DARTS-} & 0.1500 & 0.2301 & 0.7017 & 0.9259  \\

& $\Lambda$-DARTS\cite{Movahedi2022lamda} & 0.1706 & 0.2780 & 0.7471 & 0.9442 \\

& \textbf{EM-DARTS (Ours)}  & \textbf{0.1481} & \textbf{0.2202} & \textbf{0.6664} & \textbf{0.9034} \\

\bottomrule
\end{tabular}
}
\vspace{-12pt}
\label{table5}
\end{table}
\begin{table}[!htbp]
\caption{Results of various approaches on the JuDo1000 database}
\centering
\resizebox{\columnwidth}{!}{
\begin{tabular}{ p{0.9cm} c c c c c }
\toprule
\multirow{2}*{Approach} & \multirow{2}*{JuDo1000} &\multirow{2}*{EER} & \multicolumn{3}{c}{FRR@FAR}\\
\cline{4-6}
& &  & $10^{-1}$ & $10^{-2}$  & $10^{-3}$\\
\hline
\multirow{5}*{DL-based } & DEL\cite{Makowski2021DeepEyedentificationLiveOB} & 0.1238 & 0.0781  &  0.5508 & 0.8945 \\

& Dense LSTM\cite{Taha2023EyeDriveAD} & 0.0669 & 0.0195 & 0.2305 & 0.6016\\

& RP-Inception\cite{RP-InceptionV3} & 0.0743 & 0.0560 & 0.2975 & 0.6216\\

& EKYT\cite{Lohr2022EyeKY}& 0.0773 & 0.0125 & 0.1953  & 0.4922 \\

& EmMixformer\cite{Qin2024EmMixformerMT} &0.0543 & 0.0059 & 0.1284 & 0.3359 \\

\hline 

\multirow{5}*{NAS-based} & DARTS \cite{Liu2018DARTSDA} & 0.0583 & 0.0339 & 0.2035 & 0.4918\\

& Fair DARTS\cite{Chu2019FairDE} & 0.0588 & 0.0365 &0.1953  & 0.4848 \\

& DARTS-\cite{Chu2020DARTS-} & 0.0577 & 0.0357 & 0.1911 & 0.4772 \\

& $\Lambda$-DARTS\cite{Movahedi2022lamda} & 0.0412  & 0.0194  & 0.1024  & 0.2853 \\

& \textbf{EM-DARTS (Ours)}  & \textbf{0.0377} & \textbf{0.0175} & \textbf{0.0854} & \textbf{0.2422} \\

\bottomrule
\end{tabular}
}
\vspace{-12pt}
\label{table6}
\end{table}
\begin{table}[!htbp]
\caption{Results of various approaches  on the EMglasses database}
\centering
\resizebox{\columnwidth}{!}{
\begin{tabular}{ p{0.9cm} c c c c c }
\toprule
\multirow{2}*{Approach} & \multirow{2}*{EMglasses} &\multirow{2}*{EER} & \multicolumn{3}{c}{FRR@FAR}\\
\cline{4-6}
& &  & $10^{-1}$ & $10^{-2}$  & $10^{-3}$\\
\hline
\multirow{5}*{DL-based } & DEL\cite{Makowski2021DeepEyedentificationLiveOB} & 0.1853 & 0.3565  & 0.8877 & 0.9900\\

& Dense LSTM\cite{Taha2023EyeDriveAD} & 0.2069 & 0.4545 & 0.8928 & 1.0000 \\

& RP-Inception\cite{RP-InceptionV3} & 0.1743 & 0.2932 & 0.7894 & 0.9601\\

& EKYT\cite{Lohr2022EyeKY} & 0.1892 & 0.3444 & 0.8131  & 0.9812 \\

& EmMixformer\cite{Qin2024EmMixformerMT} & 0.1599 & 0.2572 & 0.7888 & 0.9711\\

\hline 

\multirow{5}*{NAS-based} & DARTS \cite{Liu2018DARTSDA} & 0.1752 & 0.2908 & 0.7597 & 0.9481\\

& Fair DARTS\cite{Chu2019FairDE} & 0.1796  & 0.3332  & 0.8022 & 0.9917 \\

& DARTS-\cite{Chu2020DARTS-} & 0.1786 & 0.3502 & 0.8122 & 0.9722 \\

& $\Lambda$-DARTS\cite{Movahedi2022lamda} & 0.1738 & 0.3324 & 0.7931 & 0.9701\\

& \textbf{EM-DARTS (Ours)}  & \textbf{0.1358} & \textbf{0.2173} & \textbf{0.7034} & \textbf{0.9348} \\

\bottomrule
\end{tabular}
}
\vspace{-12pt}
\label{table7}
\end{table}

\subsection{Ablation Experiment}

Our EM-DARTS improves the baseline DARTS in several ways. First, we proposed a local search strategy to find the optimal architecture of each cell independently,  without parameter sharing. Then, we proposed a global search strategy to optimize the target network. The optimal CNN architecture is found by performing a global and a local search in an alternate order. To improve its performance, the resulting CNN is then simplified by discarding the layers with the lowest entropy. To evaluate the performance of each component, we gradually add the local search strategy, global search strategy, and transfer entropy strategy to traditional DARTS, with the resulting models denoted as `+ Local search', `+ Global search', and `+ TES'. The experimental results on the RAN sub-dataset are reported for comparison. Verification error rates are presented in Table \ref{table10},  respectively. The results show that integrating local search, global search, and transfer entropy improves verification accuracy, which implies the importance of each component.

 {In addition, to analyze the hyper-parameters sensitivity, we conducted additional ablation experiments, the results of which are presented in Table \ref{table14}. We used EM-DARTS as a baseline and performed experiments on the JuDo1000 dataset at different search spaces, different cells, and different normalization operators.}

 {
In experiments, our EM-DARTS include 6 cells and the parameters in Table \ref{table12}. Also, EM-DARTS uses a softmax function in Eq. \ref{eq2} to relax the categorical choice of a particular operation over all possible operations. We denote it as the baseline for comparison. Then, we change the size of the convolution kernels and add new operators in the search space. Specifically, by replacing `seq\_conv\_3’ and `dil\_conv\_3’ in Table \ref{table12} with `seq\_conv\_7’ and `dil\_conv\_7’, the resulting model is denoted as ‘+ Kernel size’. In addition, we add the self-attention operation to enlarge the search space, and the resulting model is denoted as `+Self-attention’. From the experimental results in Table \ref{table14}, we can see that increasing the size of the convolutional kernel is capable of enhancing the network's receptive field, resulting in an accuracy improvement of accuracy. Meanwhile, incorporating the self-attention operation improves the representation capacity of time series features. However, to achieve a fair comparison with existing DARTS-based approaches, we employ the same search space in our experiments.  Additionally, we also show the recognition error with different numbers, e.g., 4, 5, 7, 8. The experimental results show that our searching architecture with 6 cells, namely our EM-DARTS achieves the lowest EER, e.g., 0.0377, which implies that the network architecture is further optimized by the transfer entropy. To investigate the performance of different relaxation functions, we relax the categorical choice of a particular operation to the Sigmoid function and Min-Max normalization function. The results imply that employing softmax for relaxation achieves higher performance.  Such a result can be explained by the following facts. For the Min-Max normalization function, it is difficult to process the outlier well and the Sigmoid activation function suffers from the problem of Vanishing Gradient. During backpropagation, on moving towards deep networks, the gradient becomes very close to 0. So, weight doesn't get updated much leading to very slow convergence. By contrast, Softmax shows the following advantages. First, the Softmax function converts an input vector into a probability distribution. Second, the function is differentiable, allowing for gradient-based optimization methods, which are essential for training neural networks. Thirdly, the use of the exponential function can magnify differences in the input values, assisting in distinguishing between various variables.
}

 {In our architecture search, normalization functions are critical for selecting operations by normalizing action scores. Ablation experiments comparing Sigmoid, Min-Max normalization, and Softmax (Table \ref{table14}) highlighted Softmax as the most effective choice. Unlike Sigmoid and Min-Max, Softmax maintains a stable probability distribution, ensuring balanced consideration of action scores. Its robustness to weight fluctuations prevents erratic behavior and promotes consistency during selection. Softmax’s suitability for multi-class scenarios aligns well with the multi-operation requirements of architecture search, leading to improved performance, diverse architectures, and better generalization.}

\begin{table}[!htbp]
\caption{The component ablation experimental results on the RAN}
\centering
\begin{tabular}{ c c c c c }
\toprule
\multirow{2}*{Approach} &\multirow{2}*{EER} & \multicolumn{3}{ c }{FRR@FAR}\\
\cline{3-5}
&  & $10^{-1}$ & $10^{-2}$  & $10^{-3}$\\
\hline
DARTS \cite{Liu2018DARTSDA} & 0.0956 & 0.0910 & 0.5412 & 0.8987 \\
+ Local search  & \multirow{2}*{0.0777} & \multirow{2}*{0.0599} & \multirow{2}*{0.3648} & \multirow{2}*{0.7654}\\
(without parameter sharing)  &   &   &   &  \\
{+ Global search}  & 0.0609 & 0.0370 & 0.2080 & 0.4913\\
\textbf{$+TES$ (EM-DARTS)} & \textbf{0.0537} & \textbf{0.0304} & \textbf{0.1686} & \textbf{0.4226} \\

\bottomrule
\end{tabular}
\vspace{-10pt}
\label{table10}
\end{table}

\begin{table}[!htbp]
\caption{ {The parametric ablation experimental results on the JuDo1000}}
\centering
\begin{tabular}{ c | c c c }
\toprule
\multirow{2}*{Protocol} & \multirow{2}*{Approach} & \multirow{2}*{EER} & Search Cost\\
& & & GPU days\\
\hline

Baseline & EM-DARTS & \textbf{0.037}7 & \textbf{1.45} \\
\hline

\multirow{2}*{Search space} & + Kernel size & 0.0314 & 2.51 \\
 
 & + Self-attention & 0.0291  & 3.62 \\

\hline

\multirow{4}*{Number of cells} & Four cells & 0.0796 & 0.86\\
 
 & Five cells & 0.0742 & 1.09 \\
 & Six cells (Baseline) & 0.0377 & 1.45 \\
 & Seven cells & 0.0518 & 2.12 \\

 & Eight cells & 0.0632 & 2.73\\

\hline

\multirow{2}*{Score normalization} & Sigmoid & 0.0528 & 1.45\\
 
 & Min-Max & 0.0574 & 1.46 \\
 & Softmax (Baseline) & 0.0377 & 1.45 \\

\bottomrule
\end{tabular}
\vspace{-10pt}
\label{table14}
\end{table}

\subsection{Occlusion Experiment}
 {
To evaluate the robustness and performance of EM-DARTS under occlusion or noisy conditions, we experimented on the JuDo1000 dataset. To build the occlusion testing datasets, we randomly set the eye movement data points to zero with different occlusion ratios, e.g., from 0\% to 20\% with an interval of 5\%. As a result, there are five occluded testing sets. Then we directly input the occlusion data to the training models for recognition and the experimental results of various approaches are reported for comparison.}

\begin{figure}[htbp]
\centerline{\includegraphics[scale=0.8]{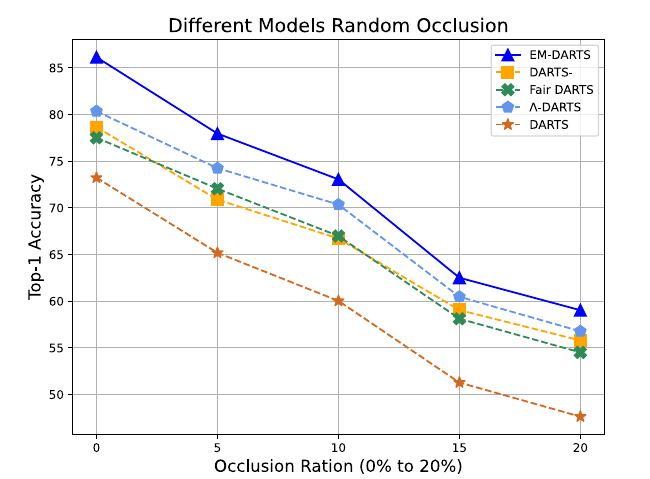}}
    \caption{ {The performance of different models at different occlusion ratios on HSS.}}
    \label{occlusion}
    \vspace{-15pt}
\end{figure}

 {
As illustrated in Fig. \ref{occlusion}, EM-DARTS consistently outperform existing models at all occlusion levels and achieves the highest accuracy. Overall, the accuracy of all approaches is degraded with increasing occlusion ratio. The results may be attributed to the following facts. The subtle variations in high-precision time-series data are very crucial for eye movement recognition. Therefore, the occlusion may change the feature distribution, resulting in the reduction of recognition accuracy.}



\begin{table*}[!htbp]
\caption{ {Efficiency comparison of different methods on RAN sub-
dataset from Gazebase\\
$\dag$: The results do not include the search phase}}
\centering
\scalebox{0.85}{
\begin{tabular}{c| c |c c c c c c c c c c}
\toprule
\textbf{Data Set} & \textbf{Metric} & \textbf{EM-DARTS$^\dag$} & \textbf{Fair DARTS$^\dag$}& \textbf{DARTS-$^\dag$}& \textbf{DARTS$^\dag$} &\textbf{$\Lambda$-DARTS$^\dag$}& \textbf{EKYT}& \textbf{Dense LSTM} & \textbf{PR-Inception}&\textbf{DEL}&\textbf{EmMixformer}\\
\midrule

\multirow{4}*{RAN} & Time (s/seq) & 0.018 & 0.017 & 0.017  & 0.017 & 0.019 & 0.092 & 0.299 & 0.941 & 0.070 & 0.249\\
    & Params (M) & 0.42  &  0.33  &  0.30 & 0.28 & 0.34 & 7.82 & 1.06 & 27.66 & 0.53 & 3.79\\
    & FLOPs (GB) & 1.47 & 1.08 & 1.15 & 1.96 & 1.17 & 0.31 & 1.38 & 6.19 & 0.17 & 1.57\\
    & Performance (EER) & 0.0537 & 0.0720 & 0.0788 & 0.0956  & 0.0726 & 0.0885 & 0.1161 & 0.0893 & 0.1436 & 0.0801\\
\bottomrule
\end{tabular}
}
\label{table13}
\vspace{-10pt}
\end{table*}
\begin{table}[!htbp]
\caption{ {Compared to the search performance of various algorithms on JuDo1000.\\
$\sharp$:Recorded on a NVIDIA A100 tensor core GPU.}}
\centering
\begin{tabular}{ c c c c }
\toprule
\multirow{2}*{Algorithm} & $Search Cost^\sharp$ & Params & \multirow{2}*{EER} \\
   & (GPU days) & (MB) &  \\
\hline
DARTS &  1.44  &  2.82  &  0.0583   \\
Fair DARTS & 1.42  & 2.75 & 0.0588 \\
DARTS- & 1.44 &  2.82 &   0.0577  \\
$\Lambda$-DARTS & 1.66  & 2.90  &   0.0412  \\
EM-DARTS & \textbf{1.45} &  \textbf{2.83} &   \textbf{0.0377}  \\

\bottomrule

\end{tabular}
\label{table1}
\vspace{-5pt}
\end{table}

\subsection{Efficiency Comparison}
 {
To evaluate the efficiency of our approach more thoroughly, we compute EER, the number of model parameters (Params), the average processing time per sequence (s/seq), and the Floating Point Operations per Second (FLOPs) of various comparable methods on the RAN sub-dataset. EER represents a performance metric, Params indicate model complexity, s/seq reflects inference efficiency, and FLOPs represent computational complexity. Table \ref{table13} reports these metrics for the five automatically designed models and five manually designed models. Also, we compare the architecture search Cost, complexity, and accuracy of all DARTS-based architecture searching algorithms on the JuDo1000 database in Table \ref{table1}. From Table \ref{table13}, we can see that the architecture searching models achieve the lowest verification error with comparable time cost, computational complexity, and the number of parameters. This can be explained by the following facts. 1) Unlike existing DARTS-based searching methods \cite{Liu2018DARTSDA}\cite{Chu2019FairDE}\cite{Movahedi2022lamda}\cite{Chu2020DARTS-} share the same architecture for all layers, our search model search different architecture for different layers, which improves the diversity of model architecture. 2) Our architecture searching method optimized the target neural network architecture in an alternate way, which avoids creating a gap between the architecture depths in the search and evaluation scenario. 3) We investigate a transfer entropy-based simplification strategy to simplify the target neural network, to determine optimal number of layer.}

 {Besides, the experimental results in Table \ref{table1} show that our searching approach achieves lower searching time cost with the lowest EER. Indeed, similar to existing DARTS-based searching models, our architecture searching method results in additional time cost, but the resulting model achieves low recognition time (e.g., 0.018 seconds) during testing. In practical application, the additional searching cost and recognition time are acceptable. Overall, our EM-DARTS achieves a balance between accuracy and efficiency on eye movement recognition.}

\subsection{Ground Truth Dataset Comparison}
 {To further evaluate the model’s generalization, robustness, and effectiveness of our preprocessing steps, we compared the model’s performance on the original (ground truth) datasets and the preprocessed datasets. In the preprocessing stage, we transformed the raw coordinate-based eye movement data into velocity-based data and segmented it into saccades (fast eye movements) and fixations (slow eye movements). As shown in Table \ref{table15}, the proposed approach achieves poor performance, e.g. high EER and low ACC on original data (coordinate) compared to the processed data (velocity). This improvement can be attributed following facts. The preprocessing approach may reduce the redundancy of law data and highlight the important patterns and thus enabling the search model to effectively learn distinctive features for eye movement recognition. }
\begin{table}[!htbp]
\caption{ {Comparative EER and ACC with respect to datasets and ground truth datasets.\\
$\ddagger$:Not preprocessed.}}
\centering
\scalebox{0.85}{
\begin{tabular}{ c | c c | c c | c c }
\toprule
Index & JuDo1000 & JuDo1000$^\ddagger$ & RAN & RAN$^\ddagger$ & EMglasses & EMglasses$^\ddagger$\\
\hline

EER & 0.0377 & 0.0708 & 0.0537 & 0.1170 & 0.1358 & 0.1554 \\

ACC(\%) & 86.23 & 64.25 & 76.40 & 55.62 & 51.98& 42.57\\

\bottomrule
\end{tabular}
}
\vspace{-10pt}
\label{table15}
\end{table}

\section{Conclusion}
In this paper, we have proposed EM-DARTS, a hierarchical differentiable architecture search algorithm to automatically learn the optimal architecture for eye movement biometric authentication. First, we have investigated a local differentiable architecture search strategy to find the optimal architecture of each cell instead of parameter sharing, which improves the diversity of the search architecture. Second, we have integrated a global search strategy to uncover the optimal architecture of the target network by alternate training. Finally, transfer entropy is proposed to simplify the CNN architecture, to reduce redundancy. Our experimental results on three representative datasets have demonstrated that our approach outperforms existing methods and achieves a new state-of-the-art verification accuracy on eye movement recognition.

 {Our approach achieves a promising performance for eye movement tasks, but it still suffers from following limits. 1) Currently, our approach includes common usage candidate operations such as CNN into search space, which may limit the representation capacity. 2) The testing datasets are relatively small, so the performance of our approach can not be sufficiently verified for large-scale eye movement recognition tasks. In the future, we will explore more operators to improve the robustness and generalization of our approach and build a large-scale eye movement database to promote the development of eye movement recognition. Our future research will focus on more efficient and generalized automatic network search methods, utilizing larger eye movement datasets to advance the field. We aim to relax search policy constraints, reduce prior knowledge's impact, and explore more efficient candidate operations in the future.}
 {In addition}, eye movement signatures can be combined with other biometric features to create a highly secure multimodal biometric system. Finally, given the intimate association between eye movement signals and EEG signals, we will further explore their potential as predictive and diagnostic indicators of neurological diseases in our future research.

 {\printnomenclature}
\hfill 

\nomenclature{EEG}{Electroencephalogram}
\nomenclature{EER}{Equal Error Rate}
\nomenclature{TPR}{True Positive Rate}
\nomenclature{FPR}{False Rejection Rate}
\nomenclature{FAR}{False Acceptance Rate}
\nomenclature{TES}{Transfer Entropy Strategy }
\nomenclature{FXS}{Gaze task}
\nomenclature{BLG}{Eye-driven game task}
\nomenclature{HSS}{Horizontal scanning task}
\nomenclature{VD}{Video viewing tasks}
\nomenclature{TEX}{Reading task}
\nomenclature{RAN}{Randomized scanning task}
\nomenclature{GD}{Gradient-based Differentiable}
\nomenclature{RL}{Reinforcement Learning}
\nomenclature{EA}{Evolutionary Algorithms}
\nomenclature{DEL}{DeepEyedentificationLive}
\nomenclature{CNN}{Convolutional Neural Network}
\nomenclature{RNN}{Recurrent Neural Network}
\nomenclature{LSTM}{Long Short Term Memory}
\nomenclature{OPC}{Oculomotor Plant characteristics}
\nomenclature{DAG}{Directed Acyclic Graph}
\nomenclature{DARTS}{Differentiable Neural Architecture Search}
\nomenclature{NAS}{Neural Architecture Search}
\nomenclature{SVM}{Support Vector Machine}
\nomenclature{PCA}{Principal Component Analysis}
\nomenclature{FDM}{Fixation Density Map}
\nomenclature{ML}{Machine Learning}
\nomenclature{DL}{Deep Learning}
\nomenclature{ROC}{Receiver Operating Characteristic}
\nomenclature{PR}{Precision vs Recall}


\bibliographystyle{unsrt}
\bibliography{TEX}

\end{document}